# A generic tool to generate a lexicon for NLP from Lexicon-Grammar tables

Matthieu Constant and Elsa Tolone
*Université Paris-Est*

Symbolic approaches to deep parsing often require large-coverage and fine-grained lexical information, such as a syntactic lexicon. Lexicon-Grammar tables (Gross 1975, 1994), carefully developed by linguists since the 70s, constitute such a syntactic resource. Each table represents a class of predicates sharing some syntactic features. Each row corresponds to a lexical entry (verb, predicative noun, predicative adjective, adverb, fixed expression) and each column corresponds to a syntactic feature (construction, argument distribution, and so on). However, they are not directly exploitable for NLP applications because pieces of information are not formally encoded although their informal descriptions are available in the literature.

Some projects such as (Hathout et Namer 1998, Gardent *et al.* 2006, Sagot et Fort 2007, Danlos et Sagot 2009) attempted to reformat Lexicon-Grammar tables in a lexicon for NLP. In these projects, each class is assigned a specific configuration which encodes missing information and defines restructuration operations. For instance, each configuration in (Gardent *et al.* 2006) is represented by a graph that makes the class structure explicit and translates each column header into a feature structure. Nevertheless, Lexicon-Grammar tables are continually updated to be improved (e.g., addition and renaming of features) and this approach can be tedious to maintain. For example, if a same feature is added to several classes, all corresponding configurations have to be modified. In this paper, we describe LGExtract, a tool that uses a global approach. First, it relies on the so-called table of classes, which encodes pieces of information that are undefined in the original classes, especially features that are constant over a whole class. Next, as a syntactic feature has exactly one interpretation over the set of classes, our extraction script assigns to each feature a set of reformatting operations once.

This paper is organized as follows. First, we briefly describe the Lexicon-Grammar classes and the table of classes, and their relevance to our work. Then, we present LGExtract in detail, illustrate it with a concrete example for French and discuss its main advantages and drawbacks.

## 1. Classes in the Lexicon-Grammar

While modern linguistics, under the generative influence, has been trying to model human language on the basis of a rather small number of samples, scholars working in the Lexicon-Grammar framework have been concentrating on the construction of syntactic and lexical databases for more than thirty years (Gross 1975, Boons *et al.* 1976, Guillet et Leclère 1992, Gross 1994). The Lexicon-Grammar methodology consists in establishing a taxonomy of syntactic-semantic classes whose lexical items share some syntactic features. For instance, class 33 contains verbs that enter the construction with one indirect complement introduced by preposition *à*. Each class is represented by a table that includes all lexical items of the class. If a verb has two meanings, it is divided into two lexical items: in the verb class 33 (see Figure 1), *se rendre* has two meanings and therefore two lexical items:

(1)     *Jean s'est rendu à mon opinion (John finally accepted my opinion)*

(2)     *Vercingetorix s'est rendu à Cesar (Vercingetorix surrendered to Caesar)*

A selection of features is applied to all entries and their linguistic validity is checked. At the intersection of a row corresponding to a lexical item and a column corresponding to a feature, the cell is set to '+' if it is valid or '-' if is not. For instance, one meaning of *se rendre* (to accept) accepts a non human nominal complement in its canonical sentence: its feature *N1 =: N-hum* value is true ('+') while it is false ('-') for the other (to surrender). There are also some features whose values are lexical items. For instance, prepositional complements can require different prepositions depending on the predicate: in class 1, which is composed of auxiliary verbs followed by a preposition and an infinitive, *arrêter* (to stop) requires preposition *de* and *commencer* (to begin) requires preposition *à*.

In the classification of French verbs, for example, there are 13,400 verb entries grouped into 60 syntactic classes. The same principles have been applied to the classification of nominal predicates, with approximately 10,300 lexical entries. Figure 2 shows a sample of a predicative noun class from (Giry-Schneider 1987). In the same way, 42,400 fixed expressions have been described.

Figure 1. *Sample of verb class 33*

| N0 =: Nhum | N0 =: N-hum | N0 =: Nnr | Ppv | Ppv =: se figé | Ppv =: en figé | Ppv =: les figé | Nég | \<ENT\> | N0 V | N0 être V-ant | N1 =: Nhum | N1 =: N-hum | N1 =: le fait Qu P | Ppv =: lui | Ppv =: y | N0hum V W sur ce point | [extrap] | \<OPT\> |
|---|---|---|---|---|---|---|---|---|---|---|---|---|---|---|---|---|---|---|
| + | - | - | \<E\> | - | - | - | - | renaître | + | + | - | + | - | + | - | - | | Max renaît au bonheur de vivre |
| + | - | - | se | + | - | - | - | rendre | + | - | + | + | + | - | + | + | | Max s'est rendu à mon opinion |
| + | - | - | se | + | - | - | - | rendre | + | - | + | + | - | - | - | - | | Le caporal s'est rendu à l'ennemi |
| + | - | - | \<E\> | - | - | - | - | renoncer | - | - | + | + | - | + | - | - | | Max renonce à son héritage |

Figure 2. *Sample of predicative noun class FNAN*

| \<ENT\>N | autre Det | Det =: un | Det =: un-Modif | Det =: du | Det =: des | N0 faire le N de V0-inf W | N0hum faire Det N à N1hum sur ce point |
|---|---|---|---|---|---|---|---|
| cadeau | \<E\> | + | + | - | + | + | - |
| calembour | \<E\> | + | + | - | + | - | + |
| câlin | \<E\> | + | + | - | + | - | - |
| canular | \<E\> | + | + | - | + | + | + |
| carambouilles | \<E\> | - | - | - | + | - | - |

## 2. Table of classes

Some basic pieces of information in the Lexicon-Grammar classification are left implicit in the current version of the Lexicon-Grammar, so they cannot be exploited by NLP tools. For instance, a feature is often explicitly recorded in the entries of a class if its value varies from one entry to another. In particular, classes are defined on the basis of features which are not explicitly recorded in the lexicon. These definitions are only described in the literature. To tackle this issue, the notion of table of classes has been defined following (Paumier 2003). Its role is to assign features to classes when possible, i.e., when their value is constant over a class (e.g., class definition features). Each row stands for a class and each column stands for a feature. Each cell corresponds to the validity of a feature in a class. Two cases can occur:

> the values depend on the entries of the class and must be assigned for each entry; the cell is then filled with the symbol 'o';

> the same value holds for the whole class and can be assigned in the cell (by '+' or '-').

For instance, the table of French verb classes currently constructed by researchers at the Institut Gaspard-Monge of Université Paris-Est Marne-la-Vallée is composed of 60 verb classes and 488 features. A sample of this table is given in Figure 3. In this table, we can see that defining features of class 33 are set: e.g., construction feature *N0 V à N1* is true ('+'). Construction feature *N0 V N1* is never valid ('-'). The non defining feature *N1 =: N-hum* is assigned 'o' because it depends on the lexical entries.

Figure 3. *Sample of the table of verb classes*

| table | N0 =: Nhum | N0 =: N-hum | N0 =: Nnr | N0 =: V1-inf W | <ENT> | Ppv =: se figé | N0 V | N0 V N1 | zone 1 | N0 V à N1 | N1 =: Nhum | N1 =: N-hum | N0 V Prep N1 V0-inf W | N0 V N1 V0-inf W | N0 V V0-inf W |
|---|---|---|---|---|---|---|---|---|---|---|---|---|---|---|---|
| V_2 | + | - | - | - | o | o | - | - | - | - | - | + | o | o | + |
| V_4 | - | - | + | + | o | - | o | + | - | - | o | o | - | - | - |
| V_31R | o | o | - | - | o | o | + | - | - | - | - | - | - | - | - |
| V_31H | + | - | - | - | o | o | + | - | - | - | - | - | - | - | - |
| V_33 | o | o | o | - | o | o | o | - | - | + | o | o | - | - | - |
| V_32H | o | - | o | - | o | o | - | + | - | - | + | - | - | - | - |

**3. Extracting an NLP lexicon with LGExtract**

Past proposals for reformatting Lexicon-Grammar tables into a lexicon for NLP consisted in making a specific setup for each class: selecting relevant features, providing information on the missing features and restructuring the data (Hathout et Namer 1998, Gardent *et al.* 2006). As the definition of the same reformatting operations can be repeated several times over the set of classes because some features occur in several classes, this approach can be tedious for encoding and maintenance.

We propose a more global approach by using (1) a unique script configuration covering all classes and (2) a table of classes to provide information undefined in the original classes. To implement this approach, we developed in Java a tool named LGExtract. It takes as input a configuration script and a table of classes. It parses this script thanks to a parser generated by the tool Tatoo (Cervelle *et al.* 2006). It outputs the set of lexical entries encoded in the classes covered by the table, formatted as described in the script. It is based on the following principles:

information is encoded in linguistic objects defined in the script. They are represented by lists and feature structures, that can be combined together; for example, objects define syntactic constituents, distributions of syntactic constituents, constructions, predicate-argument representations, lexical rules; the objects can be parametrized by the syntatic features available in the table of classes;

each feature of the table of classes is associated with a set of operations that combine linguistic objects together; for instance, when feature *N0 =: Nhum* is true for a given entry, an object defining a human noun phrase is added to the distribution of *N0* (i.e., the argument 0 of the predicate). If the feature is assigned true for a given lexical entry, the associated operations are activated.

This implies that each feature has one and only one interpretation over all classes, otherwise our tool will produce incorrect outputs.

A linguistic object is made up of lists and feature structures. An instance of such an object is defined by indicating its type, its name and its value. For example, the first instruction below instantiates a constituent (*const*) named *N-hum*, that is a non human noun phrase. These different objects can be combined together: e.g., a distribution is a set of syntactic constituents. In the last instruction below, *X0* contains the distribution of the argument 0: a human noun phrase (*Nhum*) and a non human noun phrase (*N-hum*).

```
define const N-hum [cat="NP",nothum="true"];
define const Nhum [cat="NP",hum="true"];
define const inf [cat="VP",mood="inf"];
define dist X0 [dist=(Nhum,N-hum),pos="0"]
```

As in every object-oriented programming language, an inheritance mechanism also exists. For instance, an infinitive introduced by preposition *à* (object *a_inf*) inherits the features of the object *inf* (defining an infinitive), and has a new feature indicating the presence of preposition *à*.

*define const a_inf inf[prep="à"];*

All these objects can be parametrized with the features of the table of classes. The parameters are of two types: boolean or string. For example, the code below defines a verbal predicate named *predV*. Its lemma is the value of the feature *<ENT>* (i.e., the lexical value of the entry). The code also indicates that the lexical rule 'passivization transformation with preposition *par* (by)' is encoded as the feature *[passif par]*.

*define pred predV [cat="verb",lemma="@<ENT>@"];*
*define lexicalRule passivePar {passivePar="@[passif par]@"};*

For each lexical entry, the parameters of the associated linguistic objects are established as follows. Each parameter corresponding to a feature is given a lexical or boolean value. The program first looks up the table of classes. If the feature has a constant value over the whole class the entry belongs to, the feature is assigned this value. If the feature value depends on the lexicon (feature value is 'o' for the line corresponding to the class), the program retrieves the value of the feature of the entry. For instance, the verb *aimer* (love) belongs to class 32H which contains transitive verbs with a human subject. The feature *[passif par]* is always true over this class. The two parametrized objects shown above would then become:

*define pred predV [cat="verb",lemma="aimer"];*
*define lexicalRule passivePar {passivePar="true"};*

Therefore, a piece of information coming from the table of classes has a higher priority than one coming from the class of the entry. If a contradiction occurs between the table of classes and a class, priority is given to the encoding of table of classes.

For each lexical entry, the program can then apply reformatting operations for each feature in the table of classes from these "lexicalized" objects. Operations are of one type only: *add* an object to another one. For instance, add an attribute-value pair or a list in a feature structure. The operations are independent of their order of application, i.e., they are non-destructive and do not depend on each other. For instance, when inserting an attribute-value pair *(a,v)* in a feature structure, if another value *ov* for attribute *a* already exists, the new value is the disjunction of *v* and *ov*. The operation is therefore non-destructive. Lists are actually sets because the result of two additions must be independent of their order of application. Before inserting a new element in a list, the program checks whether it exists or not. If it exists, it is not inserted. For instance, the following code indicates that, if feature *N0 =: Nnr* is true (i.e., *N0* is either a free noun phrase, an infinitive or a complementizer phrase), the program adds objects *Nhum*, *N-hum*, *inf*, *queP* and *quePsubj*[1] to the distribution of *N0* and inserts *N0* in the list of constituents:

```
prop @N0 =: Nnr@{
   add N0 in constituents;
   add Nhum in N0.dist;
   add N-hum in N0.dist;
   add inf in N0.dist;
   add queP in N0.dist;
   add quePsubj in N0.dist;
}
```

The resulting lexicon is generated in an XML format. XML elements and attributes can be defined by relating them with the linguistic objects in a script. This XML lexicon being hardly readable by a human, a compressed textual output has also been implemented (see examples in section 4).

## 4. An example of generated lexicon

Thanks to LGExtract, a French lexicon for NLP[2] has been generated from a selection of Lexicon-Grammar tables, i.e., all tables of verbs and predicative nouns[3], which are freely available under the LGPL-LR license. It is composed of 8,526 verbal entries (from 36 tables) and 4,475 nominal entries (from 30 tables). The extraction script only encodes a selection of features; some have been discarded because they are not exploitable. For instance, we discarded features involving nouns derived from verbs with no explicit information on the derivation procedure. Some features involving body part nouns were not considered relevant for the purposes of the paper. The generated lexicon is also provided under the LGPL-LR license. Each entry of the lexicon includes three sections:

> section **Lexical information** identifies the predicate (e.g., verb *se rendre*) and its lexical constraints (e.g., determiner distribution for predicative nouns, and prepositions in the constructions).

---

[1] *queP* and *quePsubj* are objects respectively defining complementizer phrases in the indicative and in the subjunctive moods.

[2] Several independent initiatives exist, such as Dicovalence (van den Eynde et Mertens 2006), Synlex (Gardent *et al.* 2006) or the Le*fff* (Sagot *et al.* 2006). Nevertheless, the first two do not include classes other than verbs and the latter sometimes lacks linguistic precision because it has been acquired semi-automatically.

[3] They can be found at the following url: *http://infolingu.univ-mlv.fr/english, Language Resources>Lexicon-Grammar>View*.

section **Arguments** indicates the nature of the arguments of the predicates: for instance, the argument *N0* of *canular* in class FNAN must be a human noun phrase.

section **Constructions** enumerates identifiers of all constructions of the predicate: e.g., passivization transformation or the construction *N0 Vsup Det N à N1* for the predicative noun *canular* (*Jean a fait un canular à Luc* = John made a joke to Luke).

The example below shows the code that is generated for the verbal entry *se rendre* (to surrender) of class 33. Argument 0 must be a human noun phrase. It enters constructions labeled *N0 V à N1* and *N0 V* that should be described in a grammar.

```
ID=V_33_129
lexical-info=[cat="verb",
        verb=[lemma="rendre",ppvse="true"]]
args=(const=[pos="0",
      dist=(comp=[cat="NP",hum="true",
            origin=(orig="N0 =: Nhum")])],
   const=[pos="1",
      dist=(comp=[cat="NP",hum="true",
            origin=(orig="N1 =: Nhum")],
         comp=[cat="NP",nothum="true",
            origin=(orig="N1 =: N-hum")],
         comp=[cat="leFaitComp",
            origin=(orig="N1 =: le fait Qu P")])])
all-constructions=[absolute=(construction="N0 V à N1",
            construction="N0 V"),
         relative=(construction="[extrap]",
            construction="Ppv =: y",
            construction="N0hum V W sur ce point")]
example=[example="Le caporal s'est rendu à l'ennemi"]
```

Below is an example of the nominal entry *canular* (joke) in the predicative noun class FNAN, the definition construction of which is *N0 Vsup Det N à N1*, where *N0* is a human noun phrase and *Vsup* is the light verb *faire* (make).

```
ID=N_fnan_29
lexical-info=[cat="noun",
         Vsup=[cat="verb",list=(value="faire")],
         noun=[notperm=[complete="canular"],noun1="canular"],
         detN=[list-det-modif=(det-modif=[det="un+une",
                                  modif="false"],
                        det-modif=[det="un+une",
                                  modif="true"],
                        det-modif=[det="des",
                                  modif="false"],
                        det-modif=[det="<E>",
                                  modif="false"])]]
args=(const=[pos="0",
        dist=(comp=[cat="NP",hum="true"])],
     const=[pos="1",
        dist=(comp=[cat="NP",hum="true"])])
all-constructions=[absolute=(construction="N0 Vsup Det N à N1",
                construction="N0 Vsup Det N",
                construction="N0 Vsup le N de V0-inf W",
                construction="N0hum  Vsup Det N à N1hum sur ce point")]
```

## 5. Discussions

The construction of the lexicon mentioned above enabled us to clearly identify practical advantages and drawbacks of our tool. Its main advantage is the use of the table of classes. In practice, all missing information is gathered in one single file instead of as many files as classes in the approach of (Gardent *et al.* 2006). In addition, it brings a more global linguistic view: before, the method to generate an NLP lexicon from the Lexicon-Grammar tables was to find the defining features of each class and make them explicit. Now, with the use of the table of classes, one can investigate whether a given feature is of interest for a given class. Some new linguistic questions within the Lexicon-Grammar framework may arise.

Moreover, the combination of LGExtract with the table of classes simplifies the maintenance of the NLP lexicon. First, all reformatting operations for each feature are encoded once in the script independently of the classes. Then, if it appears that a new feature is constant over a whole class, a '+' symbol simply needs to be added to the corresponding cell of the table of classes. The script does not need to be modified to add this information in the generated lexicon, because all reformatting operations corresponding to this feature have already been encoded.

The system requires that each feature has exactly one meaning in all classes. The use of the tool helps maintaining coherence in the table of classes. For instance, originally, the feature *zone*[4] is a text zone in several classes but with different interpretations:

> in most classes, it provides the lexical value of prepositions introducing verb complements independently of their position in the canonical construction.

> in class 38L0, it indicates the suffix to be added to the verb in order to obtain its derived noun.

> in class 35R, it gives an example of a complement.

We had to add new features so that each meaning gets one feature. In particular, prepositions have been numbered such that it makes it possible to directly identify the complements they introduce.

However, some limitations appeared clearly. It was sometimes necessary to repeat tens of similar operations over sets of features. For instance, it was necessary to create manually for all construction features linguistic objects differing solely in their label. This was due to the fact that the script does not allow for loops, functions with parameters, arrays and dynamic creation of linguistic objects.

---

[4] The feature *zone* didn't exist in the original classes. They were added during their conversion in an electronic format. At that stage, some feature names were simplified.

**6. Conclusions and future work**

In this paper, we have introduced a tool for generating NLP lexicons from Lexicon-Grammar tables named LGExtract. A table of classes is used to provide information missing in the classes: it makes explicit all implicit information underlying these classes. An extraction script associates each feature with a set of reformatting operations that are activated for each entry when the feature value is true. Applied to the Lexicon-Grammar tables for French, this tool produces a syntactic lexicon suitable for NLP applications such as parsing. The tool has also been experimented to generate a lexicon of predicative nouns. It shows that it can be used for predicates other than verbs. We plan to use it for extracting a lexicon of frozen expressions.

In the near future, we plan to convert our lexicon into a format that will allow its integration within a parser based on Tree Adjoining Grammar (TAG), namely FRMG (de La Clergerie 2005b, Thomasset et de La Clergerie 2006). This parser relies on the DyALog system (de La Clergerie 2005a). The underlying (factorized) TAG is automatically generated from a more abstract description level called a metagrammar. Indeed, such a symbolic description is able to take into account rich syntactic descriptions such as those provided by the lexicon presented in this paper.

**Summary**

Lexicon-Grammar tables constitute a large-coverage syntactic lexicon but they cannot be directly used in Natural Language Processing (NLP) applications because they sometimes rely on implicit information. In this paper, we introduce LGExtract, a generic tool for generating a syntactic lexicon for NLP from the Lexicon-Grammar tables. It is based on a global table that contains undefined information and on a unique extraction script including all operations to be performed for all tables. We also present an experiment that has been conducted to generate a new lexicon of French verbs and predicative nouns.

*Authors' address*

Matthieu Constant and Elsa Tolone
IGM, Université Paris-Est Marne-la-Vallée
5, boulevard Descartes
Champs-sur-Marne
77454 Marne-la-Vallée Cedex 2
France

{mconstan,tolone}@univ-paris-est.fr